\documentclass[conference]{IEEEtran}
\IEEEoverridecommandlockouts
\usepackage{amsfonts}%/mathbb
\usepackage{multirow}
\usepackage{amsmath}
\usepackage{algorithm}    
\usepackage{algorithmic}   
\usepackage{cite}
\usepackage{verbatim}%\iffalse
\usepackage{booktabs} %\specialrule
\usepackage{color}
\usepackage{subfigure}
\usepackage{graphicx} %\rotatebox
\usepackage[colorlinks,linkcolor=blue]{hyperref}

\def\BibTeX{{\rm B\kern-.05em{\sc i\kern-.025em b}\kern-.08em
    T\kern-.1667em\lower.7ex\hbox{E}\kern-.125emX}}

\begin{document}

\title{An Efficient Optical Flow Based Motion Detection Method for Non-stationary Scenes\\
{\footnotesize \textsuperscript{*}}
\thanks{}
}

\author{

\IEEEauthorblockN{1\textsuperscript{st} Junjie Huang}
\IEEEauthorblockA{\textit{Institute of Automation} \\
\textit{Chinese Academy of Sciences}\\
Beijing, China \\
huangjunjie2016@ia.ac.cn}
\and
\IEEEauthorblockN{2\textsuperscript{rd} Wei Zou}
\IEEEauthorblockA{\textit{Institute of Automation} \\
\textit{Chinese Academy of Sciences}\\
Beijing, China \\
wei.zou@ia.ac.cn}
\and

\IEEEauthorblockN{3\textsuperscript{rd} Zheng Zhu}
\IEEEauthorblockA{\textit{Institute of Automation} \\
\textit{Chinese Academy of Sciences}\\
Beijing, China \\
zhuzheng2014@ia.ac.cn}
\and
\IEEEauthorblockN{3\textsuperscript{rd} Jiagang Zhu}
\IEEEauthorblockA{\textit{Institute of Automation} \\
\textit{Chinese Academy of Sciences}\\
Beijing, China \\
zhujiagang2015@ia.ac.cn}

}

\maketitle

\begin{abstract}
Real-time motion detection in non-stationary scenes is a difficult task due to dynamic background, changing foreground appearance and limited computational resource. These challenges degrade the performance of the existing methods in practical applications. In this paper, an optical flow based framework is proposed to address this problem. By applying a novel strategy to utilize optical flow, we enable our method being free of model constructing, training or updating and can be performed efficiently. Besides,  adaptive intervals and adaptive thresholds are designed to heighten the system's adaptation to different situations. The experimental results show that our method adapts itself to different situations and outperforms the state-of-the-art real-time methods by 24.2\% on DAVIS2016 J-means, indicating the advantages of our optical flow based method.
\end{abstract}


\section{INTRODUCTION}

In this paper, we study the real-time motion detection which is useful in many practical applications like increasing the sensory ability in autopilot, capturing significant cues for event analysis or providing intelligence in monitoring.

Aiming at detecting motion from complex scenes, many methods have been proposed and developed in depth. They can be classified roughly into two categories: one is with real-time property, but poor performance\cite{MCD,SCBU} and the other is relatively high-performance but time-consuming \cite{MODNet,LMP,FST,FSEG}. To pursue high-performance while maintianing the real-time property, we propose an optical flow based framework which is able to efficiently detect motion. As the 2D projection of the real world motion, optical flow directly reflects the scene's moving information between two frames. This makes optical flow based method much more effective in moving foreground detection when compared to the other methods. The strategy of our efficient method is to estimate a background optical flow field $\tilde{\textbf{f}}$ and compare it with the mixed optical flow field $\textbf{f}$ in pixel-level to judge out the moving foreground. More specifically, we consider the distribution of background optical flow field as a quadratic function of the point coordinates. This is a polynomial regression estimation of the realistic complex distribution, and its accuracy is guaranteed by the small camera motion between the two consecutive frame. We sample points in the mixed opitcal flow field to perform least squares regression estimation and use our dedicated Constrained RANSAC Algotithm(CRA) to improve both the accuracy and the speed.

%is obtained by analyzing the mixed optical flow field.

\begin{figure*}[tbp]
  \centering
  \includegraphics[width=0.8\hsize]{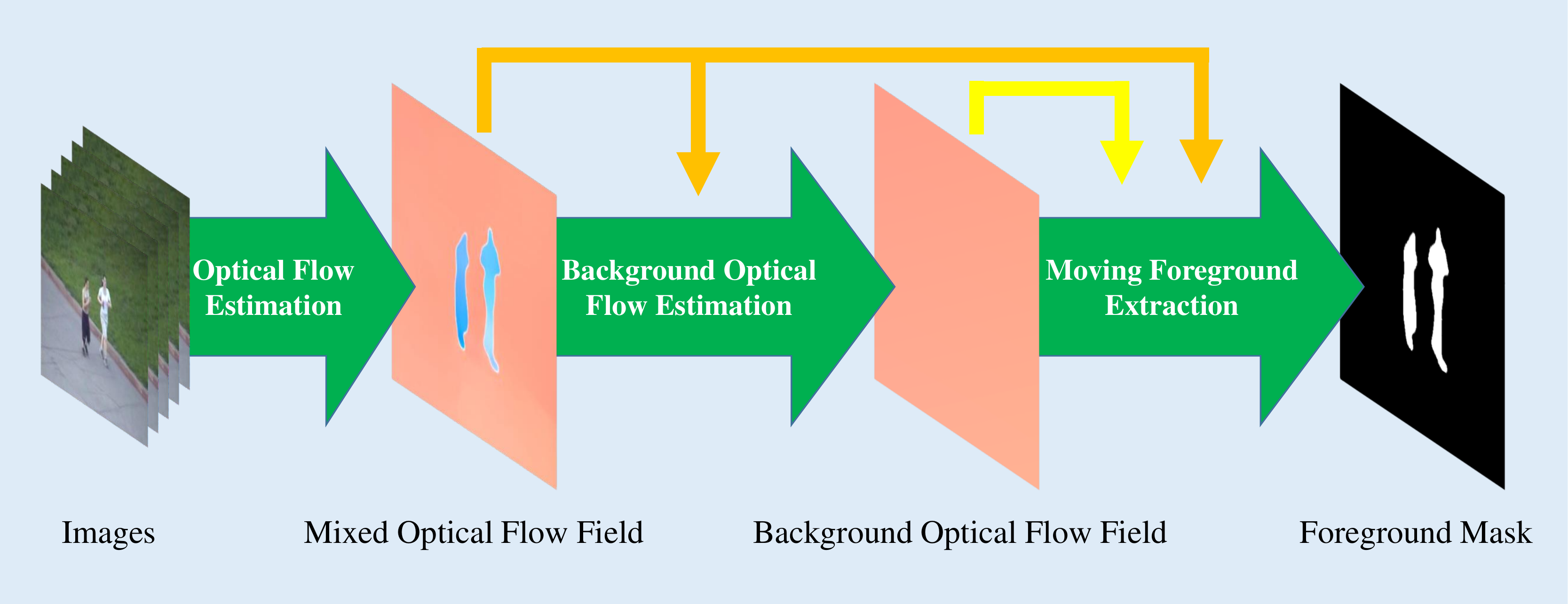}
  \caption{The proposed motion detection framework.}
  \label{fig1}
\end{figure*}

Subsequently, the moving foreground is judged out by setting a threshold for the vector difference $\textbf{f}_d = \tilde{\textbf{f}}-\textbf{f}$ between the optical flow in the mixed optical flow field and that in the background optical flow field. In practice, the spacial distribution of $\textbf{f}_d$ in background area is too complicated to apply a fixed threshold like that in \cite{DOFMOD} and many other works. Firstly, the distribution interval size of  $||\textbf{f}_d||$ in background area is linear related to the speed of the camera motion. Secondly, if there's evident zooming component in scene change, the distribution interval of $||\textbf{f}_d||$ is so large that usually have overlap with that in the foreground area. In other words, there may be no threshold value that can be used to completely separate the moving foreground from background. Unlike many other works who pay less attention on the threshold designing, we propose to use an adaptive threshold that matches the speed of scene's motion to avoid making too much false positive judgments.  

The contributions of this work are as follows: Firstly, we propose a novel and effective optical flow based motion detection framework. The framework doesn't need model constructing, training, or updating before it used and thus can be performed online. Moreover, it is also efficient enough for general real-time applications. Secondly,  adaptive invervals and adaptive thresholds is introduced to strengthen the system's adaptation to different situations.

The remainder of this paper is organized as follows. Section~\ref{sec:RW} reviews the related work. Our detection framework based on optical flow is detailedly introduced in Section~\ref{sec:MT} and its effectiveness is verified in Section~\ref{sec:EP} by comprehensive experiments. Finally, Section~\ref{sec:CC} is devoted to conclusions.

\section{Related Work}
\label{sec:RW}
From different perspectives, a long line of works have studied the problem of motion detection in non-stationary scenes. We review recent algorithms in terms of several main modules: Gaussian model based, stochastic approximation based and optical flow based.

\textbf {Gaussian Model Based.} The method proposed in \cite{MCD} used Dual-Mode Single Gaussian Model (SGM) to model the background in grid-level, and utilized homography matrixes between consecutive frames to accomplish motion compensation by mixing models. Foreground was figured out by estimating the feature's conformity to the corresponding SGM. Benefitting from Dual-Mode SGM, the method can reduce the foreground's pollution to the background models. Analogously, Yun and Jin \cite{FPSMOD}, and Kurnianggoro et.al. \cite{BSMC} used a foreground probability map and simple pixel-level background models respectively to fine-tune the result obtained in \cite{MCD}. The background models constructed and updated by these methods lack a reflection to the essence of the problem. They are sensitive to parameters and outputt result with low recall.
%and lack of robustness to different scenes. As shown in our qualitative comparison experiment result(Fig.~\ref{fig4}), for relatively big moving object, the Gaussian model based methods are valid in area near the edge but make false negative judge in the internal area of the moving objects. Besides, they also perform poor if the foreground color is similar to the background's. These make the Gaussian model based methods outputting result with low recall. 

{\bf Stochastic Approximation Based.} Francisco and Ezequiel \cite{SA} used some predefined features to form a mixture model representing the distribution of feature in previous frame. Then they achieved the motion compensation by interpolating a full covariance matrix of the pixel models. The moving foreground was judge out according to the probability of the point feature belonging to the mixture model. The performance of this method relies on the manual selected features to a large extent and is poor in the situation that the scene changes fast.

{\bf Optical Flow Based.} Kurnianggoro et.al. \cite{DOFMOD} modeled the background using zero optical flow vectors instead. After using a homography matrix to align the previous frame, dense optical flow was estimated between the alignment result and the current frame. Finally a simple optical-flow magnitude threshold was used to judge out the foreground points. As the homography matrixes are only used for aligning, the background model and the judge mechanism constructed by this method are too simple to deal with intricate unconstrained scenes. Though the recall of the experiment result was rather high, the precision was much lower than that of the SGM based methods. Manjunath Narayana et.al \cite{OFO} used optical flow orientations only to deal with the change of scene depth. This makes their method performing poorly in most tracking video sequences, where the orientations of optical flow are the same in the scenes. 

There are some other methods that do not depend on any background model. They construct the contour of foreground based on detecting large gradient points in dense optical flow field. For example, Li and Xu \cite{OSBMOD} performed mathematical morphology operations on the initial contours to obtain closed boundaries. After that the maximal contour area was selected as the area of the moving object. This simple framework can be performed easily but also limits the method to simple scenes. Papazoglou and Ferrari \cite{FST} combined the optical flow's gradient and direction to generate a better contour. Then, an efficient inside-outside maps algorithm was performed to initially figure out the foreground points, which was finally fine-tuned by global optimization. The shortcoming is that the inside-outside maps algorithm can obtain reasonable result only in simple scenes that contain a single object. Moreover, the optimization operation makes it inefficient. 

Recently, benefitting from convolutional neural network and deep learning, P. Tokmakov et.al \cite{LMP}, Mennatullah Siam et.al \cite{MODNet} and Jain S D et.al \cite{FSEG} learned motion detection and score highly on public datasets. Their methods require training process and rely seriously on the completeness of the training sample. Besides, large amount of computation keeps them away from being real-time.

\section{Methodology}
\label{sec:MT}
The framework of our online detection method for motion in dynamic scenes is shown in Figure~\ref{fig1}. There are mainly three processes: mixed optical flow field estimation, background optical flow field estimation and foreground extraction. In the following, each step of the framework is introduced in detail.

\subsection{Mixed Optical Flow Field Estimation}
\label{OE}
Taking into account speed and accuracy, FlowNet2.0 \cite{FlowNet2} is used to estimate the optical flow vectors $\textbf f_{t,t-k}={\left[ \begin{array}{cc}u&v\end{array} \right ]}^T$, which project 2D locations $\textbf p_t={\left[ \begin{array}{cc}x&y\end{array} \right ]}^T$ in frame $t$ to the locations $\textbf p_{t-k}$ in specified frame $t-k$. To improve the perception of the slow motion while maintaining the accuracy of regression analysis, we design $k$ as an adaptive interval(AI) so that the expected average norm of the background optical flow maintains a fixed magnitude $\alpha_s$. $k$ is updated by:
\vspace{-2pt}
\begin{equation}
k_{t+1}=\mathcal{L}\left (\frac{\alpha_s*k_{t}}{\frac{1}{N}\sum_{n=1}^{N}||\textbf{f}_{t,t-k,n}||_2} \right)
\end{equation}
where $\textbf{f}_{t,t-k,n} $ is the optical flow of the $N$ sample background points that used in Section~\ref{BE}. $\mathcal{L}$ is a limiting function with a upper bound of 5.

\subsection{Background Optical Flow Field Estimation}
\label{BE}
In our framework, we consider the background optical flow field as a quadratic function of the point coordinates:
\vspace{-2pt}
\begin{equation}
\label{eq1}
\tilde{\textbf f}_{t,t-k}=\textbf H_{t,t-k}\cdot \tilde{\textbf p}_t
\end{equation}
where $\textbf H_{t,t-k}$ is a $6*2$ matrix containing 6 unknown parameters and $\tilde{\textbf p}_t={\left[ \begin{array}{cccccc}x^2&y^2&xy&x&y&1\end{array} \right ]}^T$. 

We randomly sample points in the mixed optical flow field to perform least squares regression estimation(LSRE). The result is optimized by RANSAC \cite{RANSAC} algorithm to exclude the outliers. We peform RANSAC algorithm with a fixed iterations $i=50$. To pursue a high fitting degree of the estimation result to the real background optical flow field, we should sample points as more as possible and as sparesly as possible. However, sampling too much points will reduce the success rate of the RANSAC algorithm. In this work, we design a constrained sampling strategy to aviod overfitting while improving the RANSAC searching efficiency. We name it as Constrained RANSAC Algorithm(CRA). The image plane is firstly divided into square pieces with a edge length of $S$ pixels. Then a certain percentage($P$) of the pieces is selected randomly and furthermore in each of these selected pieces one point is randomly selected out to construct a set of final sample points. 
%We perform RANSAC with the sample point number $N=4$ and the iterations $iter=50$, which can provide an ideal success rate above $1-(1-(1/2)^4)^{50}=0.96$, given the assumption that the background occupies area more than half of the images. To improve the efficiency of RANSAC algorithm, the sampling points are sparsely sampled in 2D image plane. Specifically, the images are partitioned into 16 pieces and $n$ of them are randomly selected, then one point is randomly chosen inside each selected pieces.

\subsection{Foreground Extraction}\label{CB}
\label{FE}
Subsequently, based on the aforementioned two optical flow fields, we judge out the foreground points  utilizing a threshold. The real optical flow of a pecific point in background area is distributed within an interval around the estimated one. So we apply an adaptive threshold(AT) to the difference between the ideal background optical flow and the actual optical flow, and obtain a foreground mask as described in formula~(\ref{eq4}):
\vspace{-2pt}
\begin{equation}
\label{eq4}
M_{t}=\{ \textbf{p}_t\mid d_v>{T_a}\},\,\,d_v=||\textbf f_{t,t-k}-\tilde{\textbf f}_{t,t-k}||_1
\end{equation}
where $d_v$ is the 2-norm of the complement vector. The adaptive threshold is defined as:
\vspace{-2pt}
\begin{equation}
\label{eq5}
%T_a=a_1+a_2*\sqrt{{\textbf H_{t,t-k}(1,3)}^2+{\textbf H_{t,t-k}(2,3)}^2}
T_a=\alpha_1+\alpha_2*\frac{1}{N}\sum_{n=1}^{N}||\textbf{f}_{t,t-k,n}||_2
\end{equation}
where $\alpha_1$ and $\alpha_2$ are the hyper-parameters used to control the magnitude of the threshold. $\alpha_1$ is the static component part corresponding to the destabilization caused by the sensor's resolution or the optical flow's precision. $\alpha_2$ is used to introduce the dynamic component part, and we use a high threshold when the sensor moves fast. The mean norm of the background optical flow is used to reflect the speed of the camera motion.

We summarize the whole procedure in Algorithm~\ref{alg1}.
\begin{algorithm}[ht]
	\caption{Motion Detection}%算法标题  
	\label{alg1}
	\begin{algorithmic}[1]%一行一个标行号  
	\STATE\textbf{Input:} images $I_t$ and $I_{t-k}$
	\STATE estimating the mixed optical flow field $\textbf f_{t,t-k}$ utilizing $I_t$ and $I_{t-k}$;
	\STATE estimating the background optical flow field $\tilde{\textbf f}_{t,t-k}$ using CRA and LSRE;
	\STATE extracting foreground mask $M_t$ utilizing \eqref{eq4}
    \end{algorithmic}  
\end{algorithm} 

\begin{figure*}[htbp]
	\centering
	\subfigure[IM] { \label{fig:a} 
	\begin{minipage}[tb]{0.15\textwidth}
		\includegraphics[scale=0.2]{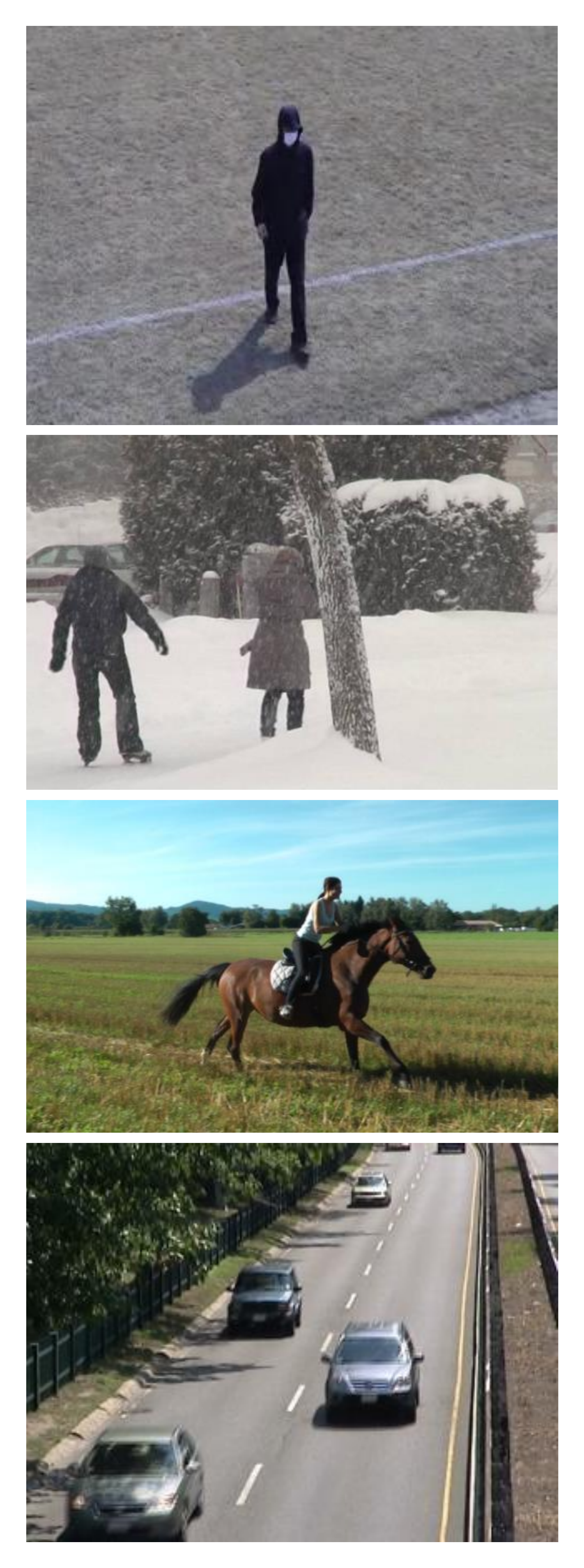}
	\end{minipage}}
	\subfigure[GT] { \label{fig:b} 
	\begin{minipage}[tb]{0.15\textwidth}
		\includegraphics[scale=0.2]{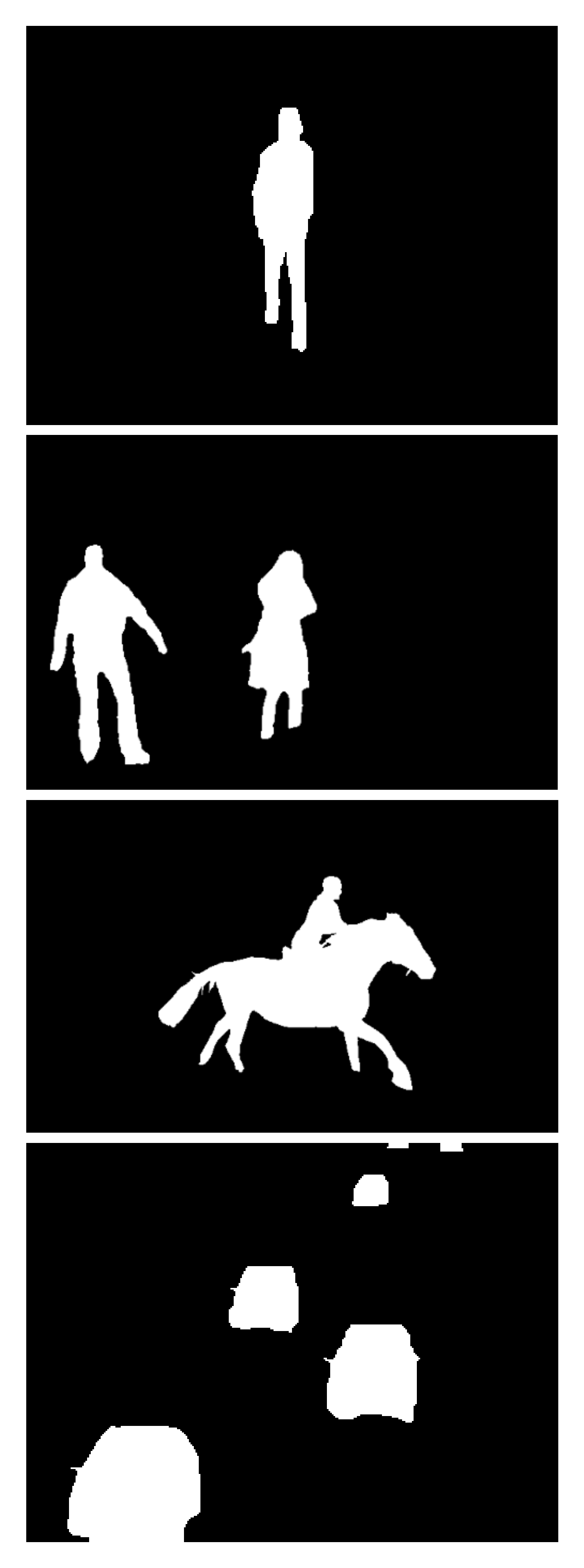}
	\end{minipage}}
	\subfigure[MCD] { \label{fig:c} 
	\begin{minipage}[tb]{0.15\textwidth}
		\includegraphics[scale=0.2]{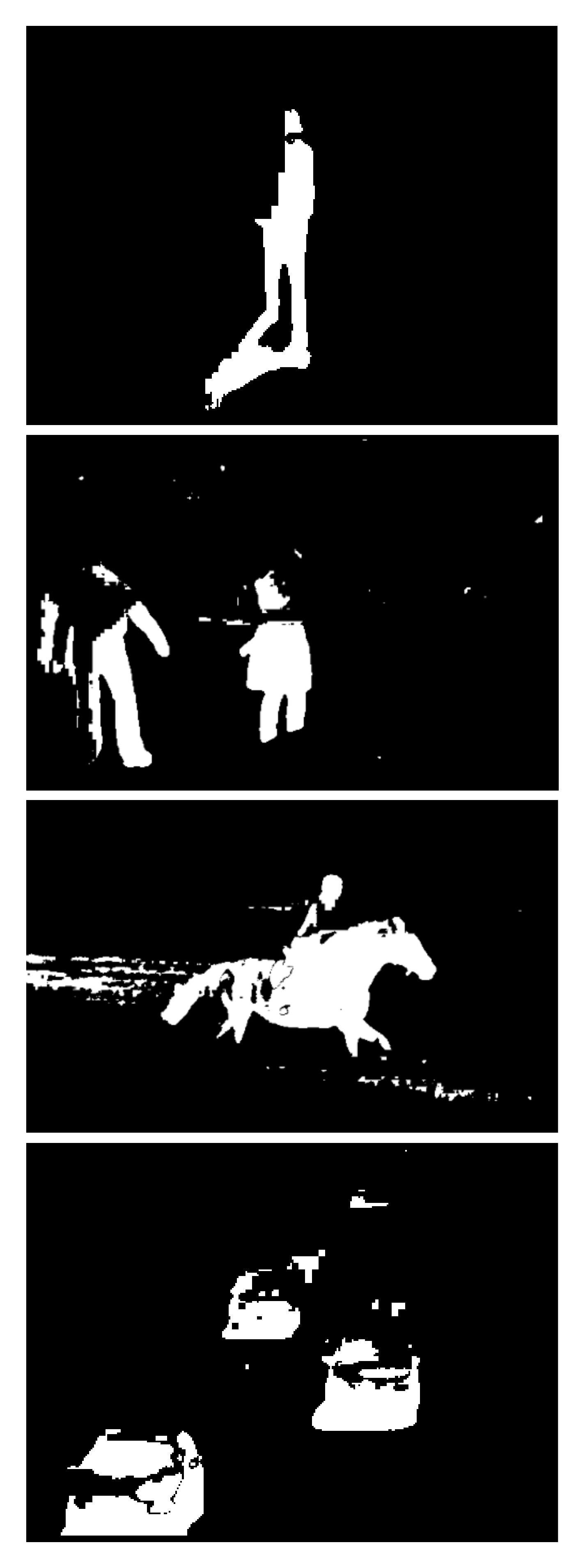}
	\end{minipage}}
	\subfigure[SA] { \label{fig:d} 
	\begin{minipage}[bt]{0.15\textwidth}
		\includegraphics[scale=0.2]{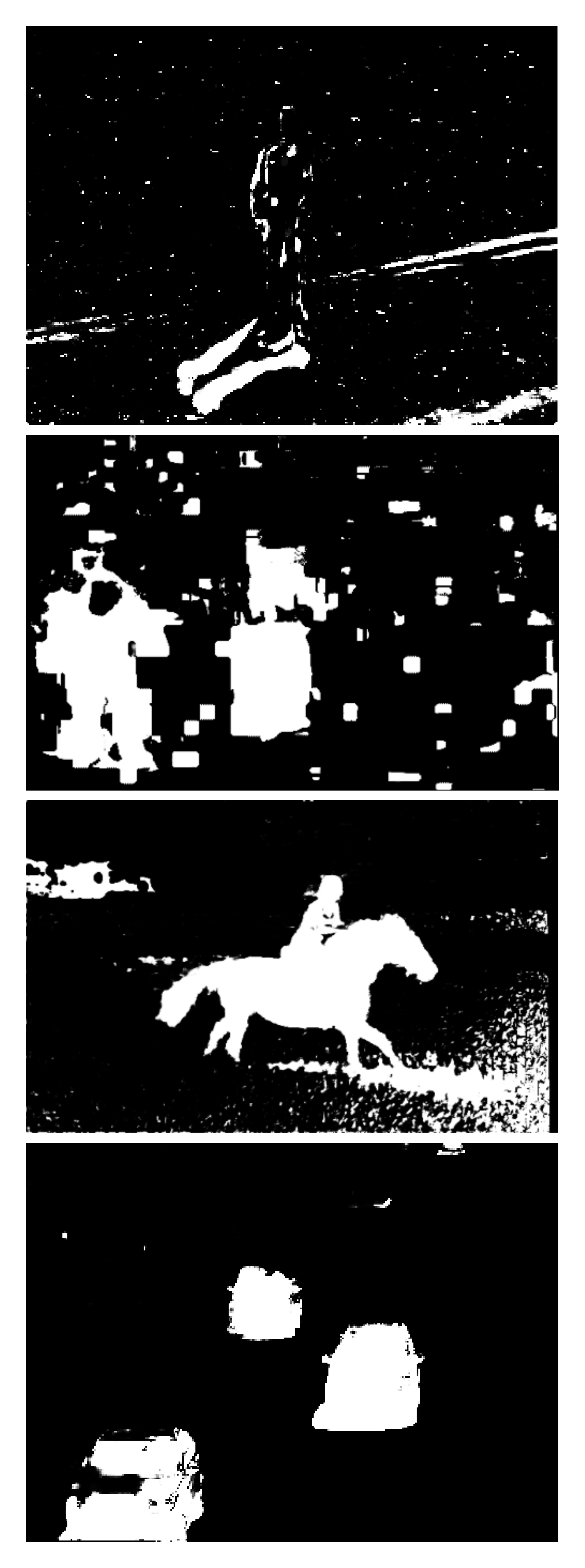}
	\end{minipage}}
	\subfigure[SCBU] { \label{fig:e} 
	\begin{minipage}[tb]{0.15\textwidth}
		\includegraphics[scale=0.2]{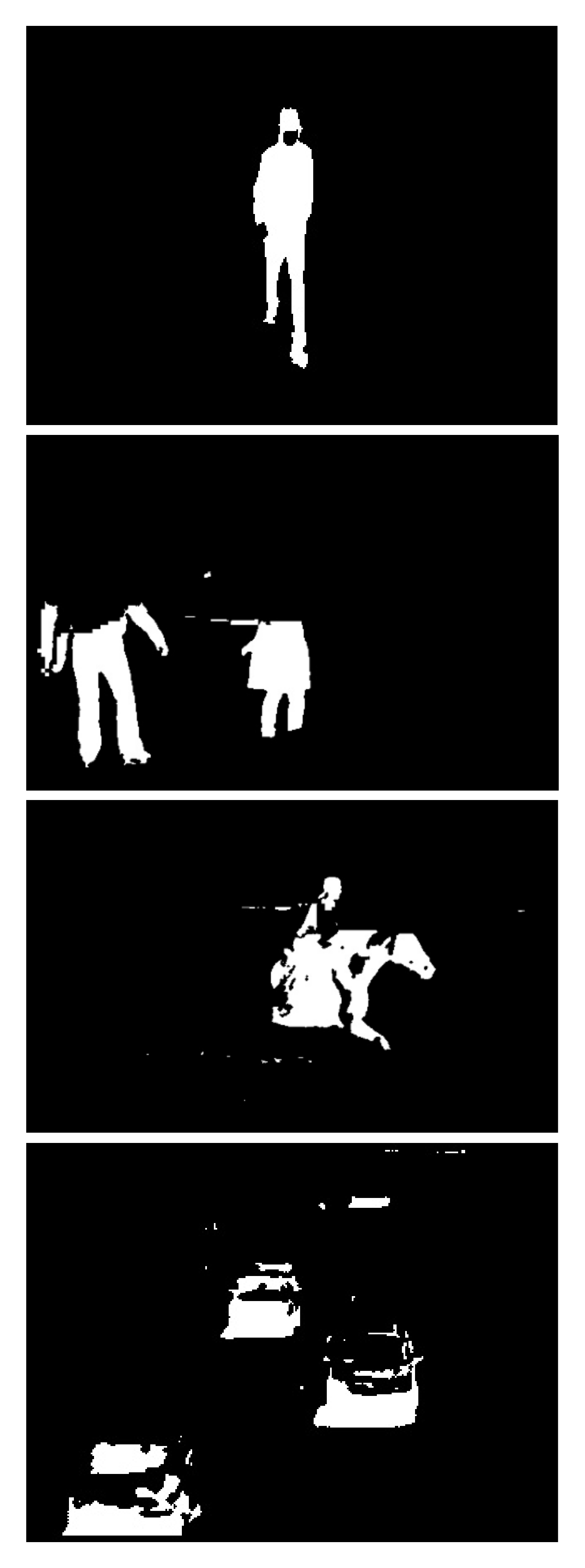}
	\end{minipage}}
	\subfigure[Ours] { \label{fig:b} 
	\begin{minipage}[tb]{0.15\textwidth}
		\includegraphics[scale=0.2]{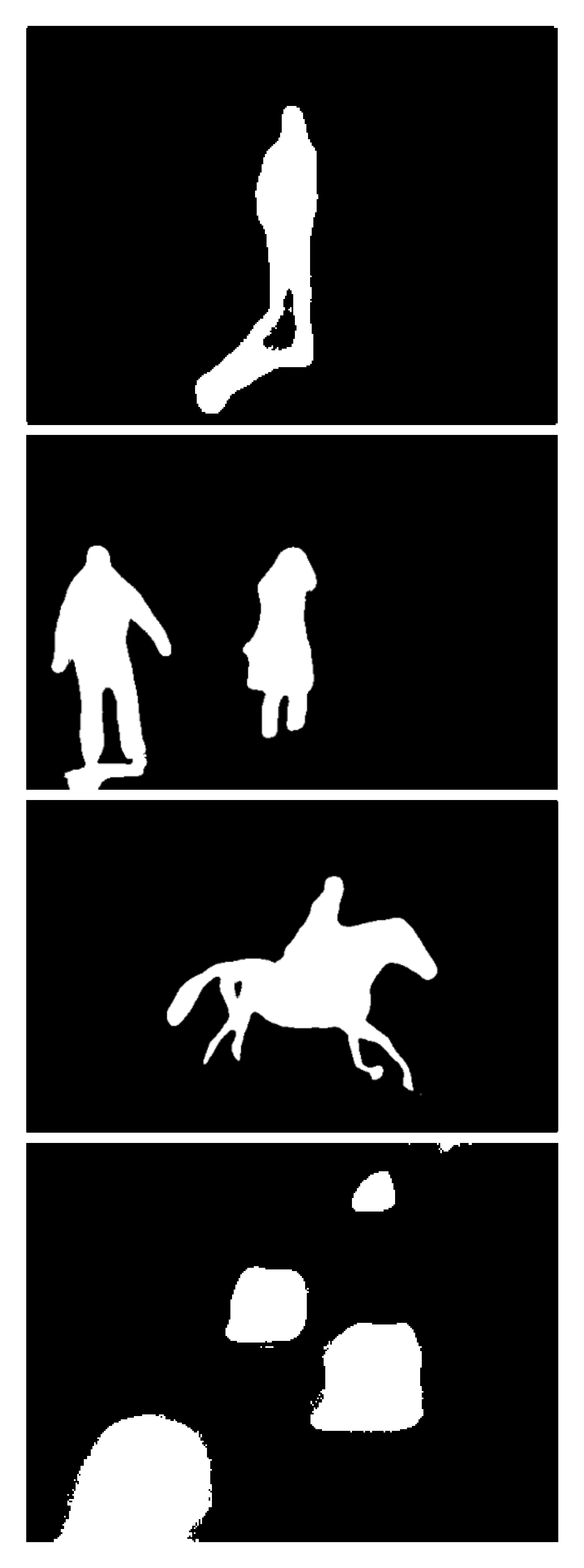}
	\end{minipage}}
	\vspace{1em}
	\caption{Qualitative results on some key frames from the additional videos. From top to bottom: Playground, Skating, Horse and Highway. The first column shows input images, and the other columns show the results of the compared methods: (a) input image(IM), (b) ground truth(GT), (c) MCD5.8ms(MCD) \cite{MCD} , (d) Stochastic Approx(SA) \cite{SA} , (e) SCBU \cite{SCBU} and (f) Our method.} 
	\label{fig4}
\end{figure*}

\section{Experiments and Results}
\label{sec:EP}
\subsection{Implementation details}
We set the parameters according to experimental regression analysis. The expected average norm of the background optical flow is set as $\alpha_s=25$. For Constrained RANSAC Algorithm, the size of pieces is set as $S=100$ and the sampling propotion is set as $P=0.5$. For foreground judging, parameters are set as following: $T_g=0.032$, $a_1=2.85$, $a_2=0.33$ and $T_c=0.99$.

\subsection{Datasets}
The proposed method is tested on the challenging DAVIS2016\cite{DAVIS} benchmark. DAVIS2016 is made up of 50 sequences, with 3455 total number of frames. It comprises a majority of challenging situations that present in motion detection, except for the situtaion with multiple objects or the situation with evident scene zooming. Thus, as supplementary, we collect an additional dataset which contains four public video sequences that contain the aforementioned missing situations. The specific information of each video is described in Table~\ref{tab1}.

\begin{table}[h]
	\centering
	\caption{Characteristics of the employed videos.}
	\label{tab1}
	\begin{tabular}{l|l|l}
		\hline
    		Video Sequences      			& NoF 	& characters \\ \hline
   	 	Playground\cite{SCBU}          	& 1466	& Evident scene zooming \\ \hline
    		Horse\cite{CDNet2014}           & 500 	& Evident scene zooming \\ \hline
    		Skating\cite{SCBU}              & 182 	& Multiple objects \\ \hline
		Highway\cite{CDNet2014} 	     	& 480 	& Multiple objects \\
    		\hline
  	\end{tabular}
\end{table}

\begin{table*}[t]
	\centering
	\caption{Quantitative comparison of different methods.}
	\label{tab2}
	\begin{tabular}{l|l||c|c|c|c|c|c|c||l|r}
		\hline
		\specialrule{0pt}{0pt}{1pt}
    		      				& \multirow{3}{*}{Method}			&\multicolumn{7}{|c||}{DAVIS}  					&Optical Flow			&Algorithm  \\ 
										&   				&\multicolumn{3}{|c|}{$\mathcal{J}$\cite{DAVIS}}&\multicolumn{3}{|c|}{$\mathcal{F}$\cite{DAVIS}}&\multicolumn{1}{|c||}{$\mathcal{T}$\cite{DAVIS}}				
																								&Estimating			&Runtime$\downarrow$ \\ 
										&         			&Mean$\uparrow$&Recall$\uparrow$&Decay$\downarrow$&Mean$\uparrow$&Recall$\uparrow$&Decay$\downarrow$&	Mean	$\downarrow$		&Method	 			& \\  \hline
   	 	\multirow{6}{*}{\rotatebox{90}{Accurate}}  & SA\cite{SA}		&28.2&17.0&-8.5&21.0&2.7&-9.0&-					&- 					&- 		\\ 
 										& FST+CRF\cite{FST}		&57.5&65.2&4.4 &53.6&57.9&6.5 &27.6				&LDOF\cite{LDOF}		&$<$3s	\\ 
										& MODNet\cite{MODNet}	&63.9&-   &-   &-   &-   &-   &-				&- 					&$>$8s 	\\ 
    		            						& MODNet+CRF\cite{MODNet}	&66.0&-   &-   &-   &-   &-   &-				&- 					&$>$8s 	\\
    		              	 					& LMP\cite{LMP}			&62.7&-   &-   &-   &-   &-   &-				&LDOF					&$>$1s	\\ 
										& LMP+CRF\cite{LMP}		&69.7&82.9&5.6 &66.3&78.3&6.7 &68.3				&LDOF					&$>$1s	\\
										& FSEG\cite{FSEG}		&\textbf{70.7}&83.5&1.5 &65.3&73.8&1.8 &32.8				&BeyondPixels\cite{BeyondPixels}					&-	\\\hline
 										
		\multirow{5}{*}{\rotatebox{90}{Fast}}      & MCD5.8ms\cite{MCD}	&29.4&13.8&-7.7&41.6&32.7&3.6&82.7				&- 					&$\approx30ms$	\\ 
 										& SCBU\cite{SCBU}	&20.5&6.0&-15.4&38.7&32.8&-9.5&112.8				&- 					&$\approx30ms$	\\ \cline{2-11}
		          	 						\specialrule{0pt}{0pt}{1pt}
										& Ours			&53.6&62.2&5.7&49.1&51.5&7.8&72.4									&FN2-css-ft-sd\cite{FlowNet2}	&51ms 	\\ 
%										& Ours+LE+FT 			&40.4&42.9&5.0&42.0&38.5&6.1&91.1				&FN2\cite{FlowNet2}			&202	\\ 
%										& Ours+QE+FT 			&49.6&&&&&&									&FN2					&202	\\
%										& Ours+QE+AT+CRA			&54.8&63.1&6.1&53.0&56.9&8.8&68.6				&FN2					&202 	\\ 
						          	 		& Ours 			&\textbf{56.1}&65.8&5.3&53.5&57.6&8.4&65.7									&FN2\cite{FlowNet2}					&202ms	\\

    		\hline
		%\specialrule{0pt}{0pt}{1pt}
		%\multicolumn{10}{l}{Note: FT denotes Fixed Threshold, AT denotes Adaptive Threshold.}
  	\end{tabular}
\end{table*}

\subsection{Qualitative results}\label{AA}
Our method is compared with the following methods for motion detection in non-stationary scenes: MCD5.8ms \cite{MCD}, SA \cite{SA} and SCBU \cite{SCBU}. Fig.~\ref{fig4} shows the qualitative results on some key frames from the additional dataset.

The qualitatively comparative results can intuitively show our proposed method's stronger adaptability to different challenges comparing with the other methods. As shown in Skating sequences, SGM base methods MCD5.8ms and SCBU perform poorly when the foreground color is slightly similar to the background. According to Playground sequences, SA can not deal with the challenges of slow motion and dynamic background. By contrast, the proposed optical flow based method can export more complete moving foreground. Meanwhile, there are few false positive point in the background area profiting from the adaptive interval and  the adaptive threshold.  According to the results of Playground and Highway sequences, our method is sensitive to shadow, which leads to some false positive results and has negative influence on the quantitative results.

\subsection{Quantitative results}\label{AA}

Our method is quantitatively compared with the state-of-the-art methods on DAVIS dataset, and the results are listed in Table~\ref{tab2}. By using FN2-css-ft-sd optical flow estimating framework, our method outperform the existing real-time methods with an improvement of $24.2\%$ on $\mathcal{J}$-mean while maintaining a time consumption of 51ms. However, the performance of the mixed optical flow field estimation form a bottleneck of our method as it only takes the mixed optical flow field as input.

\subsection{The effects of some key mechanisms}\label{AA}
Table~\ref{tab3} lists some typical combinations of different mechanisms and their $\mathcal{J}$-mean scores on DAVIS dataset. Using a linear function(LF) with fixed threshold, the proposed method just scores $40.4\%$ on DAVIS $\mathcal{J}$-mean. By contrast, applying a quadratic function(QF) improves the performance by $9.2\%$, indicating the higher fitting degree of the quadratic function to the background optical flow field.

The adaptive threshold(AT) also plays an importance role in obtaining higher quality results. By using a low threshold, the framework can detect more foreground points and score highly in recall, but it makes so much false positive judge that its precision is rather low. On the other hand, a high threshold leads to both low recall and low precision as the framework identifies too many foreground points as background points. Although there is a tradeoff, a fixed threshold(FT) causes a ceiling of the performance and lowers the robustness of the proposed method. Table~\ref{tab3} records the best performance of our proposed methods with a fixed threshold. By applying adaptive thresholds, we enable the algorithm to employ a proper threshold under specific scene situation, dramatically improving the performance of the proposed method on DAVIS $\mathcal{J}$-mean by $5.2\%$. 

The adaptive interval(AI) mechanism improves the performance to a similar degree($4.6\%$) as the adaptive threshold mechanism. This is because the adaptive threshold is designed according to the scene speed which has been constrained around a fixed value by the adaptive interval mechanism. So, respectively in two ways, two mechanisms achieve the same target that is applying a proper threshold in a specific scene. However, the constraint capacity of the adaptive interval mechanism is limited, and the adaptive threshold can not improve the perception of the slow motion as well as maintain fitting effect of the quadratic functions. By combining these two mechanisms, we can improve the performance to a higher degree($6.5\%$).

\begin{table}[h]
	\centering
	\caption{Quantitative results of different combinations.}
	\label{tab3}
	\begin{tabular}{p{50mm}|c}
		\hline
		\specialrule{0pt}{0pt}{1pt}
    		Combinations     		& $\mathcal{J}-mean$ 	 \\ \hline
		\specialrule{0pt}{0pt}{1pt}
   	 	CRA+LF+FT          	& 40.4 	 \\
		CRA+QF+FT           	& 49.6 \\
		CRA+QF+AT          	& 54.8 \\
		CRA+QF+AI           	& 54.2 \\
		CRA+QF+AT+AI       	& 56.1 \\
    		\hline
  	\end{tabular}
\end{table}

\subsection{Efficiency}\label{AA}
The proposed method is implemented using python on a PC with an Intel i5-7400 CPU, 32 GB RAM, Nvidia GTX 1080 GPU. We measure the computation time with video at a resolution of $480*854$ to evaluate the efficiency of the proposed method and other two real-time methods. The computation time of each methods is listed in the last column of Table~\ref{tab2}. Method MCD5.8ms and SCBU take up to 30ms per frame which is nearly four times that reported in their papers on account of that the time consumption of SGM based methods is linear relative to the image resolution. In contrast, the efficiency of the proposed method are relying most on the mixed optical flow estimation and the iterations of RANSAC algorithm. When using FN2 framework, the mixed optical flow field estimation process occupies nine out of ten total time consumption(186ms), and the other processes spends relatively less time(16ms). By using FN2-css-ft-sd framework which only takes 35ms to estimate the mixed optical flow field, our algorithm can be sped up to 20fps at the cost of only $2.5\%$ performance degradation on $\mathcal{J}$-mean.

\section{Conclusions}

\label{sec:CC}
%In this work, We address the challenging task of real-time moving object detection with non-stationary cameras. To judge out the moving foreground, homography matrix is estimated based on the scene's mixed optical flow field firstly. Then the background optical flow field is calculated according to the homography matrix, and is compared with the mixed optical flow field. The label of each pixel is determined based on the difference between the aforementioned two optical flow fields. Besides, a dual-mode judge mechanism with adaptive thresholds is introduced to enable the system properly dealing with different situations. In experiment part, by redefining two evaluation metrics, we enable the experiment result more properly reflecting the performance of the methods. According to our experiments, the quantitative and qualitative results obtained by our framework outperform the state-of-the-art methods indicating the advantages of our optical flow based method.

In this work, we address the challenging task of real-time motion detection in non-stationary scenes. An optical flow based framework has been presented and its outstanding performance has been demonstrated by roundly experimenting. The main efficient strategy for utilizing optical flow is to estimate a background optical flow field from the mixed optical flow field and use the background optical flow field as the judge criterion. Besides,  the adative intervals and adaptive thresholds play an important role in improving our method's robustness. Though the proposed method dramatically outperforms the existing real-time method, there is still much room for improvement when compared to the exiting accurate methods. 

%In experiment part, two evaluation metrics have been redefined to enable the experiment result more properly reflecting the performance of the methods in frame level. Given the F-Measure threshold of 0.5, we obtain a frame level high success rate of $0.92$. Though this result is outstanding among the existed methods, there is still much room for improvement in the quality of the result.

%\bibliographystyle{amsplain}%plain/unsrt/alpha/abbrv/acm/ieeetr/amsplain
%\bibliography{egbib}

\begin{thebibliography}{00}
\bibitem{MCD}Moo Yi K, Yun K, Wan Kim S, et al, Detection of moving objects with non-stationary cameras in 5.8 ms: Bringing motion detection to your mobile device, Proceedings of the IEEE Conference on Computer Vision and Pattern Recognition Workshops, 27-34, 2013.
\bibitem{SCBU}Yun K, Lim J, Choi J Y, Scene conditional background update for moving object detection in a moving camera, Pattern Recognition Letters, 88: 57-63, 2017.
\bibitem{MODNet}Siam M, Mahgoub H, Zahran M, et al, MODNet: Moving Object Detection Network with Motion and Appearance for Autonomous Driving. arXiv preprint arXiv:1709.04821, 2017.
\bibitem{LMP}Tokmakov P, Alahari K, Schmid C, Learning motion patterns in videos, IEEE Conference on Computer Vision and Pattern Recognition, 531-539, 2017.
\bibitem{FST}Papazoglou A, Ferrari V, Fast object segmentation in unconstrained video, Proceedings of the IEEE International Conference on Computer Vision, 1777-1784, 2013.
\bibitem{FSEG}Jain S D, Xiong B, Grauman K, Fusionseg: Learning to combine motion and appearance for fully automatic segmention of generic objects in videos, Proc. CVPR, 1(2), 2017.
\bibitem{DOFMOD}Kurnianggoro L, Shahbaz A, Jo K H, Dense optical flow in stabilized scenes for moving object detection from a moving camera, IEEE International Conference on Control, Automation and Systems (ICCAS), 704-708, 2016.
\bibitem{FPSMOD}Yun K, Choi J Y, Robust and fast moving object detection in a non-stationary camera via foreground probability based sampling, IEEE International Conference on Image Processing, 4897-4901, 2015.
\bibitem{BSMC}Kurnianggoro L, Yu Y, Hernandez D C, et al, Online background-subtraction with motion compensation for freely moving camera, International Conference on Intelligent Computing, Springer, Cham, 569-578, 2016.
\bibitem{SA}López-Rubio F J, López-Rubio E, Foreground detection for moving cameras with stochastic approximation, Pattern Recognition Letters, 68: 161-168, 2015.
\bibitem{OFO}Narayana M, Hanson A, Learned-Miller E, Coherent motion segmentation in moving camera videos using optical flow orientations, Proceedings of the IEEE International Conference on Computer Vision, 1577-1584, 2013.
\bibitem{OSBMOD}Li X, Xu C, Moving object detection in dynamic scenes based on optical flow and superpixels, IEEE International Conference on Robotics and Biomimetics, 84-89, 2015.
\bibitem{FlowNet2}Ilg E, Mayer N, Saikia T, et al, Flownet 2.0: Evolution of optical flow estimation with deep networks, IEEE conference on computer vision and pattern recognition, 2:6, 2017.
\bibitem{RANSAC}Fischler M A, Bolles R C, Random sample consensus: a paradigm for model fitting with applications to image analysis and automated cartography, Communications of the ACM, 24(6): 381-395, 1981.
\bibitem{DAVIS}Perazzi F, Pont-Tuset J, McWilliams B, et al, A benchmark dataset and evaluation methodology for video object segmentation, Proceedings of the IEEE Conference on Computer Vision and Pattern Recognition, 724-732, 2016.
\bibitem{CDNet2014}Wang Y, Jodoin P M, Porikli F, et al, CDnet 2014: An expanded change detection benchmark dataset, Proceedings of the IEEE Conference on Computer Vision and Pattern Recognition Workshops, 387-394, 2014.
\bibitem{LDOF}Brox T, Malik J, Large displacement optical flow: descriptor matching in variational motion estimation, IEEE transactions on pattern analysis and machine intelligence, 33(3): 500-513, 2011.
\bibitem{BeyondPixels}Liu C, Beyond pixels: exploring new representations and applications for motion analysis, Massachusetts Institute of Technology, 2009.

\end{thebibliography}

\end{document}